\documentclass[
]{ceurart}

\usepackage{listings}
\usepackage{xcolor}
\usepackage{graphicx}
\usepackage[table]{xcolor}
\definecolor{workspaces3}{HTML}{639C2E}

\begin{document}

\copyrightyear{2026}
\copyrightclause{Copyright for this paper by its authors.
  Use permitted under Creative Commons License Attribution 4.0
  International (CC BY 4.0).}

\conference{TRUST-AI: The Second European Workshop on Trustworthy AI. Organized as part of the International Joint Conference on Artificial Intelligence - IJCAI/ECAI 2026. August 2026, Bremen, Germany.}

\title{Human Preference aligned Tabular Similarity}

\author{Frederik Hoppe}[%
orcid=0009-0005-8197-6594,
email=frederik.hoppe@contact-software.com,
]
\cormark[1]

\author{Astrid Franz}

\author{Marianne Michaelis}

\author{Lars Kleinemeier}

\author{Udo G\"obel}

\address{CONTACT Software GmbH, Bremen, Germany}

\cortext[1]{Corresponding author.}

\begin{abstract}
Task-agnostic tabular embeddings are increasingly used for similarity search in real-world business systems such as Product Lifecycle Management (PLM). However, leading embedding approaches are optimized primarily for prediction tasks — not for producing human preference aligned similarity rankings. We argue that standard downstream metrics are insufficient to fully assess embedding trustworthiness for similarity search and that human preference aligned evaluation is a necessary and currently missing component. We present a concrete evaluation procedure and illustrate the problem through a PLM use case.
\end{abstract}


\maketitle

\section{Introduction}

Tabular data is a fundamental and highly important data format in industry. Especially in PLM systems, a large amount of business data is stored in tabular databases. Industrial tables contain, in addition to the classical categorical and numerical data types, a variety of other modalities such as temporal data, text, or even documents. A core operation in such systems is similarity search: finding parts, requirements, or change requests that are semantically related to a given anchor item. However, keyword- and parameter-based search algorithms are dominant in many industrial applications.

In recent years, a variety of tabular representation learning methods have been developed, e.g., TableVectorizer~\cite{skrub}, TabSTAR~\cite{arazi2025tabstar}, TabPFN~\cite{hollmann2025tabpfn}, TabICL~\cite{qu2025tabicl}, Mitra~\cite{zhang2025mitra}, Limix~\cite{zhang2025limix}, and ConTextTab~\cite{spinaci2025contexttab}. These methods represent table records as vectors in a high-dimensional vector space. Thus, they allow for (semantic) search via a distance-based nearest neighbor search. Ideally, two similar records have a smaller distance than two different ones. Here, a first major problem arises: \emph{When are two records similar?} The embeddings are created with respect to a corresponding task, e.g., classification or regression. However, if a model is trained to produce embeddings for solving a classification task, it is unclear how it performs for other tasks, e.g. regression or similarity search.

Task-agnostic tabular embeddings are a potential solution for this need, as they promise flexible reuse across multiple downstream applications. However, a validation strategy for task-agnostic embedding is missing. Recent work~\cite{hoppe2025comparing,vogel2026towards} evaluates such embedding models on downstream task metrics for classification and regression tasks. However, such objective metrics are not suitable for \emph{similarity search} as they lack human-perceived similarity. Methods and metrics from Information Retrieval~\cite{harman1993trec} may help incorporate human feedback in the validation process.

However, human annotations are highly user-centered. In particular, the assessment of similarity in a PLM system strongly depends on the user preference. For example, an engineer may call two parts similar if they have the same geometric properties, a production worker would refer to similarity in terms of the same production way, and for a purchaser two parts may be similar if they come from the same supplier.

We argue that this aspect is a trustworthiness concern: practitioners who deploy task-agnostic embeddings for similarity search have no reliable way to assess whether the retrieved nearest neighbors are meaningful to domain experts. This position paper outlines the problem, proposes a human preference aligned approach, and calls on the community to adopt broader evaluation standards aligned with trustworthy AI principles~\cite{eu2019ethics,jobin2019ethics}, such as transparency and accountability.

\section{The Problem: Evaluating Task-Agnostic Tabular Embeddings for Similarity Search}

Despite significant progress in tabular representation learning, there remains a substantial gap in evaluating embeddings for similarity-based use cases, especially concerning the critical task of generating ranked lists~\cite{he2008learningtorank}. Current benchmark suites assess embeddings through downstream prediction tasks such as classification and regression. Although these metrics are objective and reproducible, they do not completely answer the question that matters most for the similarity search:
\emph{Do the nearest neighbors in embedding space correspond to what a domain expert would consider similar?}

As an example, consider a PLM system where a product engineer searches for similar past change requests. Their need extends beyond superficial table structure; they seek deep semantic connections like the same functional area, component, or problem type, along with relevant categorical attributes (priority, status, product line). This challenge is closely related to Entity Resolution or Record Linking~\cite{brizan2006survey} at a conceptual level, aiming to identify truly congruent or related entities. 
An embedding model that scores well on classification benchmarks may still retrieve change requests that are superficially similar in terms of table structure but semantically meaningless to the engineer. This problem is invisible to automated metrics and can silently undermine user trust in the AI system retrieving similar change requests. 
Furthermore, despite enormous pretraining, tabular foundation models like TabPFN still exhibit fairness bias~\cite{schiffman2025tabPFNfairness}. This highlights the critical need for diverse human preference evaluation, ensuring the system performs equitably and satisfies the diverse relevance criteria of different user groups. 

Similarity search in tabular data is related to Information Retrieval (IR) in documents~\cite{harman1993trec} which uses metrics like Mean Reciprocal Rank or Normalized Discounted Cumulative Gain to evaluate ranked lists. These measures, even when based on human relevance judgments, do not adequately consider that different persons may have different preferences. Human alignment is increasingly addressed via Reinforcement Learning from Human Feedback (RLHF) \cite{christiano2023deepreinforcementlearning} and Preference Learning \cite{furnkranz2010preference}, which optimize ranking and generative outputs to match human judgments. However, RLHF is primarily designed for sequential decision-making or generative text outputs, whereas tabular similarity is a representation learning problem of structured records. Preference Learning in IR (such as learning-to-rank from click logs) relies on homogeneous document features and massive interaction histories. In contrast, tabular datasets are highly heterogeneous and schema-specific, meaning that preferences learned on one schema cannot easily generalize to another under severe data sparsity. Under these constraints, human preference alignment can best be achieved through Interactive IR systems~\cite{borlund2019interactive}, where continuous user engagement and feedback drive the evolution and refinement of retrieval quality.

\section{Human Preference Alignment}

To address the above problem, we propose a workflow for human preference ranking of embedding similarity in tabular data. This workflow consists of three steps:
\begin{enumerate}
    \item Ingestion step: Tabular records are embedded using a pre-selected embedding model, and indexed in a vector store. The configuration used for ingestion, such as embedding model and distance function (e.g. cosine distance), is stored as metadata.
    \item Anchor selection: In the user interface, an anchor record is randomly drawn from a predefined probability measure, which may be based on the principles of uncertainty sampling or diversity sampling~\cite{he2014uncertainty}. The anchor could also be selected in a user-specific way, based on attribute filtering~\cite{li2025filtering} or hybrid retrieval~\cite{myung2025hybrid}, or even be chosen directly by the user, e.g., with respect to business criticality. The system retrieves its nearest neighbors in the embedding space based on the predefined distance function. 
    \item User assessment: The anchor record and a neighboring record are displayed side by side with their relevant fields. Users assign a pairwise similarity label (e.g., “identical”, “similar”, “slightly similar”, or “not similar”), which is then persistently stored alongside anonymized user and session metadata. To reduce label noise arising from ambiguity, inter-annotator agreement metrics, such as Cohen's $\kappa$ \cite{gwet2014stat}, are monitored across users, and disagreement cases are flagged for subsequent analysis. This process is further informed by frameworks for trustworthy annotation~\cite{ouattara2025trustwothiness}.
\end{enumerate}

Despite its simplicity, this type of workflow is an efficient tool for collecting human preference rankings as a basis for active learning \cite{cohn1996active,settles2010active}. The resulting sets of labeled record pairs per user are a valuable basis for automated quantitative evaluation of different embedding algorithms. For a selected embedding algorithm, the collected rankings can reveal important properties of the corresponding embedding space. For example, it can be checked whether nearest neighbors are consistently judged as semantically similar, or otherwise which differences between user groups exist. Statistical hypothesis testing can be implemented to compare annotations from different users.

Measures like MRR or NDCG can be computed for each user group separately, allowing a user group-specific ranking of embedding algorithms or configurations. Hence, final algorithm selection can be user group-specific, e.g., an optimal embedding algorithm for an engineer may select other table features than an algorithm that is optimized for a purchaser.

Recent advances in Entity Resolution \cite{fu2025clustering,tang2026unlocking,dasanaike2026emsemblelink} underscore the critical distinction between entity congruence and semantic similarity. Our proposed annotation framework provides a crucial refinement for Entity Resolution pipelines. It delineates when highly similar entities exhibit functional similarity or divergence, based on the designated user group. Hence, the collected labels offer valuable training data for developing robust reranking models.

It is important to note that approximate nearest neighbor (ANN) search algorithms, such as Hierarchical Navigable Small World (HNSW) \cite{HNSW}, are commonly employed to scale retrieval to large datasets. While ANN methods offer significant computational advantages over exact nearest neighbor search, they introduce a trade-off between retrieval speed and recall accuracy. To mitigate the effect of ANN noise on the collected labels, exact nearest neighbor search can optionally be used for smaller datasets, or ANN hyperparameters can be tuned to maximize recall prior to the annotation session.

\section{Aspects of Trustworthy AI}

Human preference aligned tabular similarity has several implications for trustworthy AI.
\begin{itemize}
\item \emph{Fairness:}
The proposed human preference ranking enables the development of fair AI treating user groups equitably and avoiding bias.
\item \emph{Transparency:}
The proposed workflow makes otherwise opaque neighborhood structures visible. The users can see which records are retrieved as nearest neighbors, what distance in embedding space they have, and how they compare at the content level. This does not fully explain the embedding model, but it improves transparency at the system behavior level.
\item \emph{Robustness:}
Pairwise similarity labels can support data quality improvement. They may reveal duplicates, inconsistent records, ambiguous categories, or clusters that mix unrelated cases. In this sense, human preference ranking can contribute to data quality and hence pave the way for robust AI applications which are highly dependent on data quality.
\item \emph{Accountability:}
Persistent annotation records can support internal governance, quality assurance, and external accountability. In regulated or high-impact contexts, such documentation is increasingly important.
\end{itemize}

However, human preference ranking is resource-intensive, especially when aiming to capture the diverse preferences of different user groups. Hence, the optimal balance between objective automated downstream assessment and human-centered assessment of embedding quality with respect to similarity search needs to be explored. First, how much annotation is sufficient, particularly when considering the need to represent varied user perspectives (Q1)? Guidance is needed on the amount and distribution of human ranking required to establish confidence in an embedding procedure across different user groups. Second, how should disagreement between annotators be handled, especially when such disagreements stem from legitimate differences in preference across user groups (Q2)? Semantic similarity is often graded and context-dependent. Inter-annotator agreement metrics can be useful, but they may not fully capture these group-specific legitimate differences in perspective. Third, how can bias in embedding neighborhoods be detected, and specifically how might this bias differentially affect various user groups (Q3)? Embedding models may systematically group or separate records in ways that reflect social, linguistic, or organizational biases. Human validation workflows should support the discovery of such patterns, ensuring equitable performance across all target audiences. Fourth, how can human preference ranking results feed back into system improvement, specifically to accommodate the diverse needs of different user groups (Q4)? Pairwise labels may be used to compare embedding models, tune preprocessing strategies, or, critically, to train task-specific or user-group-specific similarity models that address distinct preferences and contexts.

\section{Pilot Study: Similarity of Tickets}

\begin{figure}[b]
\centering\includegraphics[width=0.48\linewidth]{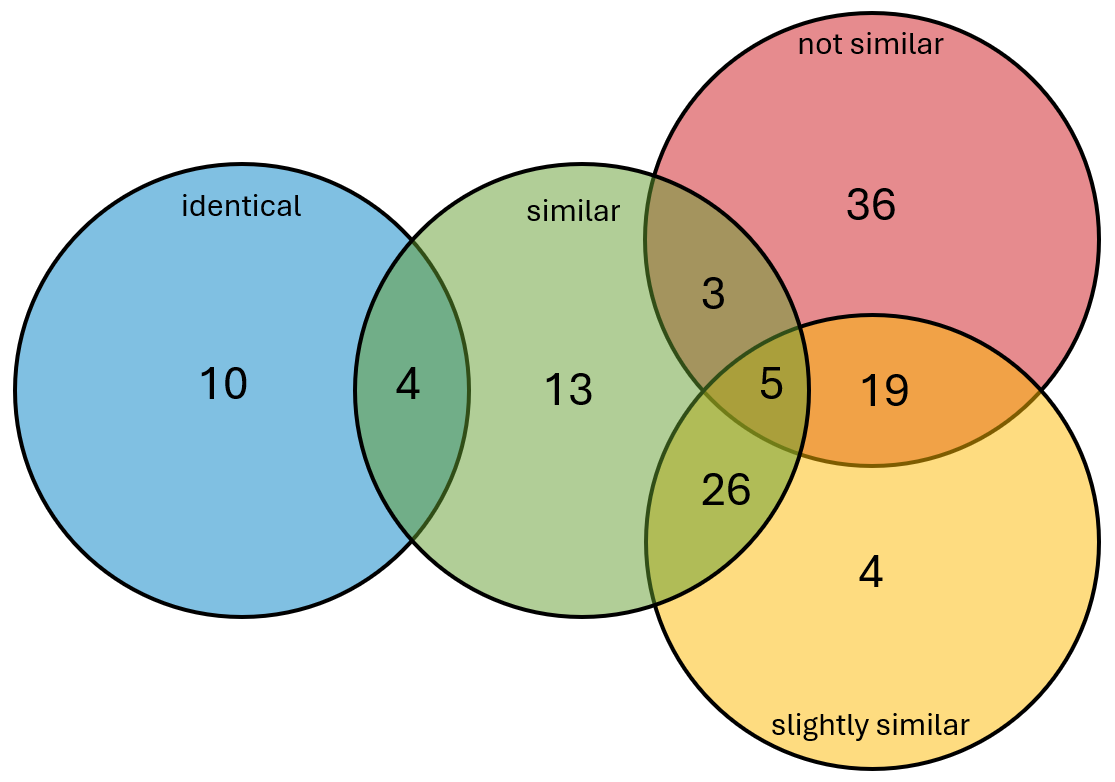}
\caption{Distribution of similarity ratings across three annotators. High rates of divergent ratings (47.5\%) illustrate the challenge of handling legitimate semantic disagreements across different user perspectives.\label{fig:diagram}}
\end{figure}

As a pilot study, we applied an embedding algorithm (Algorithm 1) to a table containing tickets from different categories (e.g., IT, safety, permission), representing an object class in a PLM system. Three different annotators rated the similarity of the top-6 nearest neighbors for 20 predefined anchor tickets on a 4-point scale (identical, similar, slightly similar, not similar), yielding 120 ticket pairs, each with three annotations. Figure~\ref{fig:diagram} illustrates the distribution of these annotations: only 63 pairs (52.5\%) received identical ratings, 52 pairs (43.3\%) were rated in two different categories, and even 5 pairs (4.2\%) were rated in three different categories. This high level of divergence grounds our second question Q2 (handling disagreement): these discrepancies are not merely random noise but likely reflect legitimate, domain-specific differences in how different users interpret \emph{similarity}. Furthermore, this setup immediately highlights a critical methodological challenge related to Q3 (bias detection): Since the candidate pairs were selected exclusively from the neighborhood of Algorithm~1, the entire evaluation dataset is biased toward Algorithm~1’s representation space. This illustrates the need for bias-aware validation workflows.

To evaluate embedding performance, we constructed annotator-specific evaluation sets of triplets $(a, x, y)$ consisting of an anchor $a$, and two comparison tickets $x$ and $y$, where the rated similarity of $(a,x)$ is strictly greater than $(a,y)$. Due to annotator disagreements, the resulting evaluation sets differed in size: 136 triplets for Annotator~1, 165 for Annotator~2, and 141 for Annotator~3. We measured the performance of 10 different embedding algorithms $e_i$ ($i=1,\ldots,10$) using triplet accuracy~\cite{schroff2015facenet,vogel2026towards}, which represents the fraction of triplets where the cosine distance $d(e_i(a),e_i(x))<d(e_i(a),e_i(y))$. The results are shown in Table~\ref{tab:accuracy}. Notably, Algorithm~7 performed best for Annotators~1 and 3, whereas Algorithm~4 was superior for Annotator~2. However, the performance of Algorithms 4 and 7 for Annotator~2 differed by only a single triplet. This marginal difference directly highlights Q1 (annotation sufficiency): with a small sample size of 120 pairs, we lack the statistical confidence to prove that Algorithm~4 is genuinely better for Annotator~2, demonstrating the need for guidelines on how much annotation is sufficient to confidently identify group-specific preference trends. Finally, these pilot results ground Q4 (system improvement) by showing how preference feedback can be operationalized: the collected triplet preferences can be used either to guide group-specific model selection (e.g., deploying Algorithm~4 for Annotator~2's group) or to optimize a single, highly robust global model like Algorithm~7 across all users.

\begin{table}
    \begin{tabular}{|c||c|c|c|c|c|c|c|c|c|c|}
    \hline
    & \rotatebox{90}{Algorithm 1} 
    & \rotatebox{90}{Algorithm 2}
    & \rotatebox{90}{Algorithm 3}
    & \rotatebox{90}{Algorithm 4}
    & \rotatebox{90}{Algorithm 5}
    & \rotatebox{90}{Algorithm 6}
    & \rotatebox{90}{Algorithm 7}
    & \rotatebox{90}{Algorithm 8}
    & \rotatebox{90}{Algorithm 9}
    & \rotatebox{90}{Algorithm 10$\,$} \\
    \hline
    \hline
    Annotator 1 & 89.7\% & 87.5\% & 88.2\% & 91.9\% & 91.9\% & 91.2\% & \cellcolor{workspaces3} 93.4\% & \cellcolor{workspaces3} 93.4\% & 87.5\% & 87.5\% \\
    (136) & (122) & (119) & (120) & (125) & (125) & (124) & \cellcolor{workspaces3} (127) & \cellcolor{workspaces3} (127) & (119) & (119) \\
    \hline
    Annotator 2 & 83.0\% & 83.0\% & 79.4\% & \cellcolor{workspaces3} 86.1\% & 84.2\% & 80.0\% & 85.5\% & 82.4\% & 78.8\% & 78.8\% \\
    (165) & (137) & (137) & (131) & \cellcolor{workspaces3} (142) & (139) & (132) & (141) & (136) & (130) & (130) \\
    \hline
    Annotator 3 & 85.1\% & 85.1\% & 81.6\% & 83.0\% & 83.7\% & 85.1\% & \cellcolor{workspaces3} 85.8\% & 85.1\% & 82.3\% & 82.3\% \\
    (141) & (120) & (120) & (115) & (117) & (118) & (120) & \cellcolor{workspaces3} (121) & (120) & (116) & (116) \\
    \hline
    \end{tabular}
    \caption{Triplet accuracy for each of the three annotators and ten different embedding algorithms. The number of triplets is given in parentheses. The best value per annotator is marked green. The tight performance gap between Algorithms 4 and 7 for Annotator 2 highlights the need for statistical sufficiency metrics and grounds the decision-making process for group-specific feedback loops.\label{tab:accuracy}}
\end{table}

\section{Conclusion}

Embedding-based similarity has rapidly become a hidden infrastructure within modern AI systems, enabling crucial functionalities such as efficient information retrieval, effective data clustering, personalized recommendation algorithms, and the generation of contextually rich input vectors for subsequent downstream models. Yet, the fundamental assumption that vector proximity reliably captures domain-relevant semantic similarity often lacks rigorous examination and comprehensive empirical validation.
In this position paper, we argued that trustworthy AI requires human-centered and auditable validation of embedding similarity with a particular emphasis on integrating diverse user perspectives. We proposed a workflow for nearest-neighbor exploration and pairwise similarity annotation in tabular data explicitly designed for user group-specific evaluation.
This work underscores a broader claim: the trustworthiness of AI systems is not solely an intrinsic property of the embedding models themselves. Instead, it is an emergent property of robust socio-technical processes through which model behavior is continuously inspected, challenged, documented, and refined. Consequently, we contend that user group-specific embedding validation must be integrated as a fundamental component of the AI lifecycle and risk management frameworks.

\section*{Declaration on Generative AI}
 During the preparation of this work, the authors used gemini-2.5-flash for rephrasing in order to improve the writing style. After using this tool, the authors reviewed and edited the content as needed and take full responsibility for the publication’s content.

\bibliography{sample-ceur}



\end{document}